\newcommand{\CLASSINPUTbottomtextmargin}{2.7cm}
\documentclass[10pt,conference,letterpaper]{IEEEtran}
\usepackage{amsmath,amsfonts,amssymb,amsthm}
\usepackage{array}
\usepackage{makecell}
\usepackage{cite}
\usepackage{float}
\usepackage{multirow}
\usepackage{booktabs}
\usepackage{float}
\usepackage{enumitem}
\usepackage{changepage}
\usepackage{amsthm}
\usepackage{graphicx}
\usepackage{xcolor}
\usepackage{balance}
\usepackage{hyperref}
\usepackage{enumitem}
\hypersetup{
    colorlinks=true,
    citecolor=red,
}
\usepackage{algorithm}
\usepackage{algpseudocode}

\theoremstyle{definition}
\newtheorem{definition}{Definition}

\theoremstyle{remark} % 设置为备注风格，通常是不斜体
\newtheorem{remark}{Remark} % 定义 remark 环境

\newcommand{\thickhline}{\noalign{\hrule height 1pt}}

\begin{document}

\title{Symmetry-Preserving Architecture for Multi-NUMA Environments (SPANE): A Deep Reinforcement Learning Approach for Dynamic VM Scheduling}

\author{
    \IEEEauthorblockN{Chan Tin Ping, 
    Yunlong Cheng, 
    Yizhan Zhu,
    Xiaofeng Gao\IEEEauthorrefmark{1}, 
    Guihai Chen}
    \IEEEauthorblockA{
        Department of Computer Science and Engineering, Shanghai Jiao Tong University, Shanghai, China.\\
        \{chantp, aweftr, zyz518348, gaoxiaofeng, chen-gh\}@sjtu.edu.cn
    }
}

\maketitle

\begingroup
\renewcommand\thefootnote{}
\footnotetext{© 2025 IEEE. Personal use of this material is permitted. Permission from IEEE must be obtained for all other uses, in any current or future media, including reprinting/republishing this material for advertising or promotional purposes, creating new collective works, for resale or redistribution to servers or lists, or reuse of any copyrighted component of this work in other works. This paper has been accepted to IEEE INFOCOM 2025.}
\footnotetext{\IEEEauthorrefmark{1}Corresponding author: Xiaofeng Gao (gaoxiaofeng@sjtu.edu.cn).}
\endgroup

\begin{abstract}
As cloud computing continues to evolve, the adoption of multi-NUMA (Non-Uniform Memory Access) architecture by cloud service providers has introduced new challenges in virtual machine (VM) scheduling. To address these challenges and more accurately reflect the complexities faced by modern cloud environments, we introduce the Dynamic VM Allocation problem in Multi-NUMA PM (DVAMP). We formally define both offline and online versions of DVAMP as mixed-integer linear programming problems, providing a rigorous mathematical foundation for analysis. A tight performance bound for greedy online algorithms is derived, offering insights into the worst-case optimality gap as a function of the number of physical machines and VM lifetime variability. To address the challenges posed by DVAMP, we propose SPANE (Symmetry-Preserving Architecture for Multi-NUMA Environments), a novel deep reinforcement learning approach that exploits the problem's inherent symmetries. SPANE produces invariant results under arbitrary permutations of physical machine states, enhancing learning efficiency and solution quality. Extensive experiments conducted on the Huawei-East-1 dataset demonstrate that SPANE outperforms existing baselines, reducing average VM wait time by 45\%. Our work contributes to the field of cloud resource management by providing both theoretical insights and practical solutions for VM scheduling in multi-NUMA environments, addressing a critical gap in the literature and offering improved performance for real-world cloud systems.
\end{abstract}


%% main text
\section{Introduction}

Cloud computing has revolutionized the way computational resources are delivered and consumed, offering unprecedented flexibility, scalability, and cost-effectiveness. 
Virtual machines (VM), the fundamental computing unit to the cloud services, enable cloud service providers (CSPs) to deliver scalable and flexible computing resources on demand. VMs are software implementations of physical machines (PM), each capable of hosting different applications and operating systems. Effectively scheduling VMs to PMs becomes a more and more crucial problem in the cloud due to the dynamic behavior of VMs, the growing number of requests and the need to guarantee Quality of Service (QoS) to users. 

\begin{figure}[tb]
\centering
\includegraphics[width=\columnwidth]{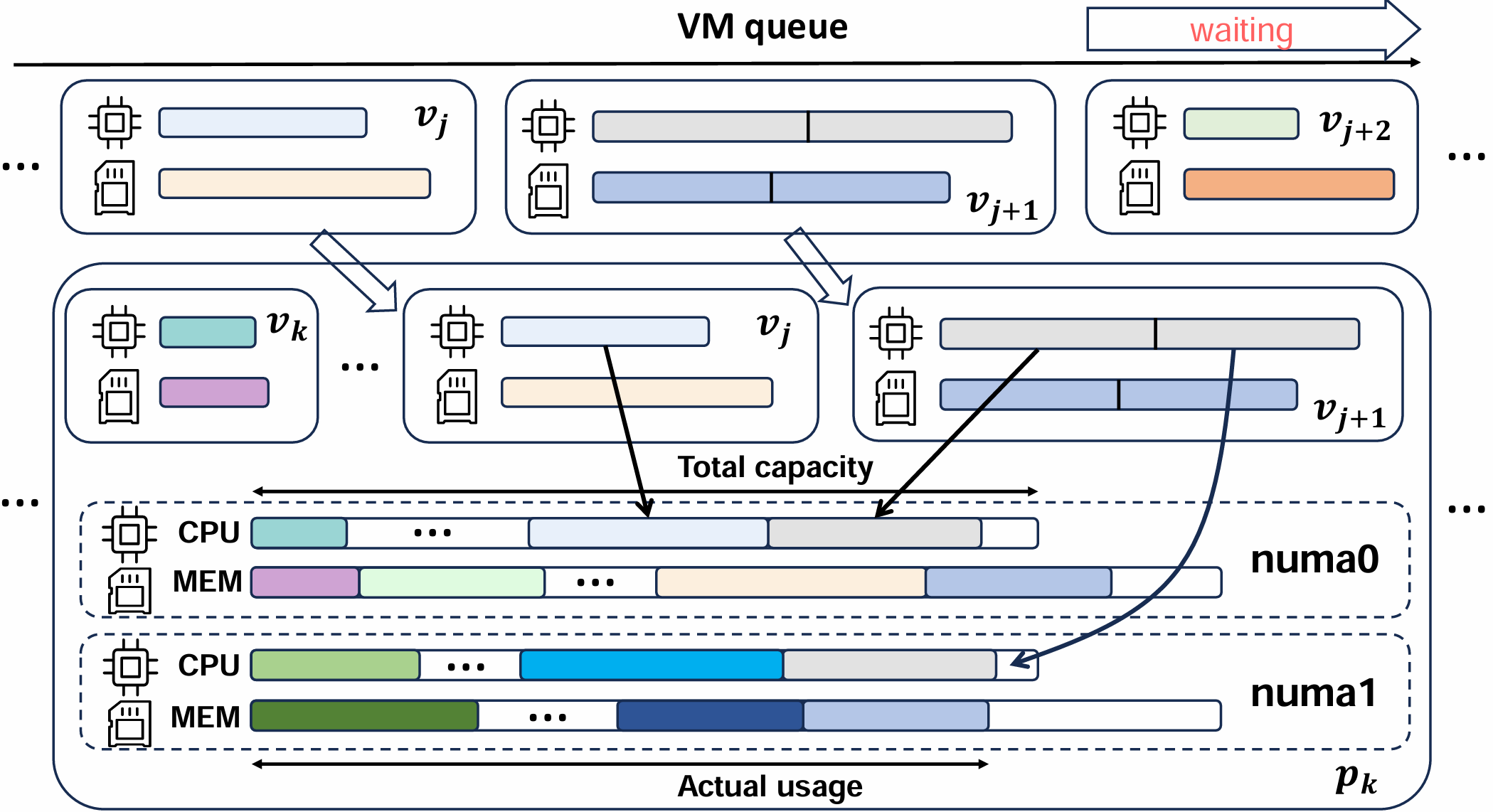}
\caption{Illustration of Multi-NUMA VM scheduling. Each PM consists of two NUMA nodes, with VMs deployed in one of three scenarios: (1) allocated entirely to a single NUMA node (e.g., $v_j$ on NUMA 0), (2) evenly distributed across both NUMA nodes within the same PM (e.g., $v_{j+1}$), or (3) delayed due to insufficient resources (e.g., $v_{j+2}$).}
\label{fig:intro}
\end{figure}

Traditionally, the VM scheduling problem is modeled as a multi-dimensional vector bin-packing problem \cite{ShiLDJS18,murhekar2023dynamic}. Given a set of VMs with varying sizes, the goal is to minimize the number of PMs needed to accommodate these VMs without exceeding capacity limits. VMs are treated as items and PMs as bins, with constraints on multiple resources such as CPU, memory, disk, and network. Many classical optimization methods, including greedy and online algorithms, have been proposed to solve this problem \cite{chan2009dynamic,murhekar2023dynamic,tang2016first,dynamicWithpre-2023}.

Multi-Non-Uniform Memory Access (NUMA) architecture, widely adopted in modern servers, improves system performance by reducing memory access latency through closer proximity of memory to processors \cite{lameter2013numa}. Many servers, such as AWS EC2 c5.24xlarge \cite{aws_numa}, feature two NUMA nodes. Compared to servers without NUMA, it offers more placement options, such as allocating VMs to a single NUMA node or splitting them across nodes, increasing the combinatorial complexity of scheduling. Fig.~\ref{fig:intro} illustrates the deployment of small and large VMs to PMs with two NUMA nodes: small VMs can be allocated to a single NUMA node, while large VMs may span multiple nodes depending on resource availability \cite{RaoWZX13multiNuma,sheng_learning_2022}. When PMs lack sufficient capacity, VM requests must wait for resources, creating a variant of the bin-packing problem. Unlike the static nature of bin-packing, VM requests in real-world cloud environments arrive sequentially and have finite lifespans, making the problem dynamic and challenging to solve.

To better capture the challenges faced by CSPs, we define the \textbf{D}ynamic \textbf{V}M \textbf{A}llocation problem in \textbf{M}ulti-\textbf{N}UMA \textbf{P}M (\textbf{DVAMP}) and present both offline and online variants. In DVAMP, the number of PMs is fixed, and VM requests must be allocated to a single NUMA node or distributed across both NUMA nodes. When immediate deployment is impossible due to resource constraints, VM requests are queued until resources become available. The primary goal of DVAMP is to minimize total VM wait time. We also provide a performance bound for online algorithms, establishing a lower bound for the worst-case optimality gap between them and the optimal.

Traditional heuristics, while commonly used, typically operate within a predefined set of rules, limiting their ability to generalize across varying cloud conditions and workloads. Moreover, in dynamic environments like cloud computing, optimal decisions are not immediately obvious. Deep Reinforcement Learning (DRL), by contrast, can continuously adapt to these changing conditions by learning from past experiences and receiving real-time feedback. This adaptability allows DRL to effectively handle the dynamic nature of VM scheduling \cite{zhou2024deep}. However, the large state and action spaces in cloud environments make both training and inference computationally expensive \cite{wang2021energy}.

In response to these challenges, we propose \textbf{SPANE}, a \textbf{S}ymmetry-\textbf{P}reserving \textbf{A}rchitecture designed specifically for Multi-\textbf{N}UMA \textbf{E}nvironments.
SPANE exploits the intrinsic symmetry of the DVAMP problem, where the ordering of hosts in the state space does not affect the scheduling outcome. By leveraging this symmetry, SPANE improves action prediction efficiency and accuracy at each scheduling step.

Extensive experiments using real-world cloud traces validate the effectiveness of SPANE, which reduces average VM wait time by 45\% compared to state-of-the-art methods. SPANE also demonstrates significant flexibility, applying effectively to VM scheduling scenarios with varying numbers of PMs. This ensures its applicability to practical cloud environments with highly dynamic and complex resource demands.

\textbf{Contributions.} We make the following contributions:
\begin{enumerate}[label=\arabic*), left=0pt]
    \item A formal definition of the DVAMP problem, including both offline and online versions. The offline version is formulated as a mixed-integer linear programming problem (MILP), establishing a rigorous mathematical foundation for detailed analysis and effective solution development. Section \ref{sec:formulation} provides the full details of this formulation.

    \item A derivation of a tight performance bound for greedy online algorithms in the DVAMP problem. This analysis shows the worst-case optimality gap is approximately $\frac{m-1}{2m-1}(\mu-1)$ times the total time-resource of VM requests, where $m$ is the number of physical machines and $\mu$ represents VM lifetime variability. This result offers insights into the limitations of online scheduling strategies and clarifies the impact of VM lifetime variability on system performance. Details are provided in Section \ref{sec:bounds}.
    
    \item The development of SPANE, a novel deep neural network architecture that exploits the inherent symmetries of the DVAMP problem. SPANE produces invariant results under arbitrary permutations of PM states, enhancing learning efficiency and solution quality. The architecture and its implementation are discussed in Section \ref{sec:RL}.
    
    \item The practical effectiveness of the proposed approach is demonstrated through extensive experiments on the Huawei-East-1 dataset~\cite{sheng_vmagent_2022}. The experimental setup, results, and analysis are detailed in Section \ref{sec:experiment}. Code and data are publicly available.\footnote{\url{https://github.com/ChanTinPing/SPANE}}
\end{enumerate}

\section{Problem Statement and Math Formulation}
\label{sec:formulation}
This section presents a formal formulation of DVAMP problem, encompassing both offline and online variants. We first introduce the system model, followed by detailed formulations of the offline and online problems.

\subsection{System Model}
\begin{definition}[VM Requests]
Let $V = \{v_1, v_2, \ldots, v_n\}$ denote the set of VM requests, where $v_j$ represents the $j$-th request. Each VM request is characterized by three essential attributes:

1) Arrival time $at_j$: The time at which $v_j$ is received by the system. Notice that VMs are processed in sequential order, which means $at_j < at_{j+1}$ for all $j$.

2) Active time length $lt_j$: The duration for which $v_j$ needs to be operational. The term "dynamic" in the problem nomenclature refers to the termination of VMs during the scheduling process.

3) Resource requirements $r^d_j$: The amount of each resource type $d$ required by the VM, where $d \in [D]$ and $D$ represents the total number of resource types (e.g., CPU cores, memory).
\end{definition}

\begin{definition}[Server Infrastructure]
The server infrastructure consists of $m$ PMs, each equipped with two NUMA nodes. Each NUMA node has the same resource capacity limit $R_d$ for each resource type $d \in [D]$. The total resource utilization of all VMs on a NUMA node must not exceed its capacity limit $R_d$ for any resource type. VM deployment adheres to one of two strategies:

1) Allocation to a single NUMA node within the same PM: 
   The chosen node's resource utilization increases by the full resource requirement of the VM.

2) Equitable distribution across both NUMA nodes within the same PM: Both NUMA nodes experience an increase in resource utilization equal to half of the VM's resource needs.

The deployment strategy for $v_j$ is determined by the variable $div_j$, defined as:
\begin{equation}
div_j = 
\begin{cases} 
1, & \text{if } r^{d_{div}}_j \geq c_{div} \\ 
0, & \text{otherwise}
\end{cases}
\end{equation}
where 1 indicates equitable distribution across both NUMA nodes. This approach balances performance and resource uilization: smaller VMs benefit from single-node allocation, boosting performance, while distributing larger VMs across nodes enables the cluster to host more VMs overall.
\end{definition}

\begin{table}[tb]
    \centering
    \caption{Symbols and Their Meanings}
    \begin{tabular}{|>{\centering\arraybackslash}m{1.9cm}|m{6cm}|} 
        \hline
        \textbf{Symbols} & \makecell[c]{\textbf{Meaning}} \\
        \hline
        [$i$] & $\{1,2, \cdots, i\}$ \\
        \hline
        $D$ & Dimension of resources \\
        \hline
        $V$ & Total VM request set \\
        \hline
        $v_{j}$ & The $j$-th VM request \\
        \hline
        $n$ & Length of total VM request set $V$ \\
        \hline
        $r^d_j$ & $d$-th resource requirement of $v_j$ \\
        \hline
        $at_j$ & Arrival time of request $v_j$ \\
        \hline
        $lt_j$ & Length of the time duration when $v_j$ active \\
        \hline
        $div_j$ & Function that determines whether $v_j$ needs to be divided; output $\in \{0,1\}$  \\
        \hline
        $p_{k}$ & $k$-th PM \\
        \hline
        $m$ & Number of PMs \\
        \hline
        $u^d_{k,i}(t)$ & The utilization of the $d$-th resource in the $i$-th NUMA of $p_k$ at time $t$ \\
        \hline
        $R_d$ & The maximum capacity of $d$-th resource of all NUMA nodes \\
        \hline
        $x_{k,i,j}$ & Whether $i$-th NUMA of $p_k$ hosts $v_j$ \\
        \hline
        $\delta_j(t)$ & Whether $v_j$ is active at time $t$ \\
        \hline
    \end{tabular}
    \label{tab:symbols}
\end{table}

\subsection{Offline DVAMP Problem}

\begin{definition}[Offline DVAMP Problem]
Given a set of VM requests $V = \{v_1, v_2, \ldots, v_n\}$, number of PMs $m$, and maximum resource capacity $R_d$ for each resource type $d \in [D]$, the offline DVAMP Problem aims to:
\begin{enumerate}
    \item Determine the optimal NUMA node(s) deployment for each VM request.
    \item Ensure that resource utilization of all NUMA nodes does not exceed $R_d$ at any time.
    \item Minimize the total wait time across all VM deployments.
\end{enumerate}
The problem is subject to potential deployment delays.
\end{definition}

\begin{remark}
In this context, "wait time" refers to the interval between a VM's arrival time and its start time, occurring when deployment is postponed (queuing time).
\end{remark}

The complete mathematical formulation of the offline DVAMP problem is presented in~\eqref{eq:DMVMS}. Table~\ref{tab:symbols} provides a comprehensive list of symbols and their corresponding definitions used in this formulation.

In the DVAMP problem, the VM wait time is quantified as the difference between its start and arrival time. The objective function in~\eqref{eq:WT} aims to minimize the total wait time across all VMs. The output of the problem, as indicated in~\eqref{eq:output}, consists of the deployment decision variable $x_{k,i,j}$ and the start time $st_j$ for each VM. Here, $x_{k,i,j}=1$ if $v_j$ is deployed on the $i$-th NUMA node of $p_k$, and $x_{k,i,j}=0$ otherwise.

\vspace{-\baselineskip} % 减少间距
\begin{adjustwidth}{-5em}{}
\begin{subequations}
\label{eq:DMVMS}
\begin{align}
& \text{Minimize: } \nonumber \\
& \qquad WT = \sum_{j=1}^{n} (st_j - at_j) \label{eq:WT} \\
& \text{Output: } \nonumber \\
& \forall k \in [m], \forall i \in \{0,1\}, \forall j \in [n]:  \nonumber \\
& \qquad x_{k,i,j}, \ st_j \label{eq:output} \\
& \text{Constraints:} \nonumber \\
& \forall k \in [m], \forall i \in \{0, 1\}, \forall d \in [D], \forall t>0: \nonumber \\
& \qquad u^d_{k,i}(t) \leq R_d \label{eq:u1} \\
& \qquad u^d_{k,i}(t) =  \sum_{j=1}^{n} \delta_j(t) x_{k,i,j} \gamma_{j} r^d_j \label{eq:u2} \\
& \qquad \gamma_{j} = 1 - \frac{div_j}{2}  \label{eq:u3} \\
& \qquad \delta_{j}(t) = \begin{cases}
1, & st_j \leq t < st_j + lt_j \\
0, & \text{otherwise}
\end{cases} \label{eq:u4} \\
& \forall k \in [m], \forall i \in \{0, 1\}, \forall j \in [n] : \nonumber \\
& \qquad x_{k,i,j} \in \{0,1\} \label{eq:x1} \\
& \qquad x_{k,0,j} = x_{k,1,j} \quad \text{if} \ div_j=1 \label{eq:x2} \\
& \forall j \in [n]: \nonumber \\
& \qquad \sum_{k=1}^m \sum_{i=0}^1 x_{k,i,j} = 1 + div_j \label{eq:x3} \\
& \forall j \in [n]: \nonumber \\
& \qquad st_{j-1} < st_j \label{eq:st1} \\
& \qquad at_j \leq st_j \label{eq:st2}
\end{align}
\end{subequations}
\end{adjustwidth}

\textbf{NUMA Resource Constraints:} The core constraint~\eqref{eq:u1} ensures that at any given moment, the resource utilization of all NUMA nodes does not exceed the maximum capacity $R_d$. Equations~\eqref{eq:u2} through~\eqref{eq:u4} delineate the calculation of the resource utilization $u^d_{k,i}(t)$. The variable $\delta_{j}(t)$ indicates whether $v_j$ is active at time $t$, as defined in~\eqref{eq:u4}. Consequently, $\delta_{j}(t) x_{k,i,j}$ determines whether $v_j$ is active in the $i$-th NUMA node of $p_k$ at time $t$. For each active $v_j$, its contribution to the resource utilization $u^d_{k,i}(t)$ is determined by $\gamma_j r^d_j$, where $\gamma_j$ is defined in~\eqref{eq:u3}. This coefficient accounts for the deployment strategy: $\gamma_j = 1$ if $div_j = 0$ (single NUMA deployment), and $\gamma_j = 0.5$ if $div_j = 1$ (equitable distribution across both NUMA nodes). Although these four equations are presented separately to enhance clarity, they can be consolidated into a single equation.

\textbf{Deployment Decision Constraints:} Constraints~\eqref{eq:x1}--\eqref{eq:x3} govern the deployment decision variable $x_{k,i,j}$.~\eqref{eq:x1} defines $x_{k,i,j}$ as a binary variable.~\eqref{eq:x2} ensures consistency for VMs that are evenly distributed across two NUMA nodes.~\eqref{eq:x3} guarantees that each VM is deployed to either one NUMA node or two NUMA nodes, depending on the value of $div_j$.

\textbf{Timing Constraints:} Finally, constraints~\eqref{eq:st1}--\eqref{eq:st2} enforce the proper ordering and timing of VM deployments, ensuring that VMs are processed sequentially and that no VM starts before its arrival time.

\subsection{Online DVAMP Problem}
In reality, cloud servers handling VM requests operate without prior knowledge of future VM arrivals, their specifications, or the termination times of active VMs. Consequently, the online version of DVAMP presents a more valuable yet challenging scenario. The mathematical formulation of the online problem is presented in~\eqref{eq:online_problem}.

\begin{definition}[Online DVAMP Problem]
The online DVAMP problem is characterized by:
\begin{enumerate}
    \item Objective: Minimize the total wait time $WT$, as defined in the offline version~\eqref{eq:WT}.
    \item Decision-making: Decisions are made sequentially as each new VM request arrives.
    \item Available Information: At time $t$, when deciding for $v_j$, only information about past and current events is known. This includes the data from~\eqref{eq:res_know} to~\eqref{eq:prev_x}, encompassing resource requirements, arrival times, start times, completed VM durations, and previous deployment decisions.
    \item Constraints: The problem is subject to the same constraints as the offline version.
\end{enumerate}
\end{definition}

\begin{remark}
This online problem mirrors real-world scenarios, requiring decisions with incomplete future information to balance current resource use and future demands. For each incoming VM request, the algorithm determines the NUMA node(s) and start time, as shown in~\eqref{eq:x_online} and~\eqref{eq:st_online}.
\end{remark}

\vspace{-\baselineskip} % 减少间距
\begin{subequations}
\label{eq:online_problem}
\begin{align}
& \text{Minimize } WT \text{ as defined in \eqref{eq:WT}} \nonumber \\
& \text{For each } j \in [n] \text{ and } t > at_j: \nonumber \\
& \quad \text{Given:} \nonumber \\
& \qquad r^d_l \qquad\qquad\quad\hspace{0.2em} \forall l \in [j], \forall d \in [D] \label{eq:res_know} \\
& \qquad at_l \qquad\qquad\quad\hspace{-0.05em} \forall l \in [j] \label{eq:at_know} \\
& \qquad st_l \qquad\qquad\quad \forall l \in [j-1] \label{eq:st_know} \\
& \qquad lt_l \qquad\qquad\quad\hspace{0.2em} \forall l \in [j-1] \ \text{s.t. } st_l + lt_l \leq t \label{eq:et_know} \\
& \qquad x_{k,i,l} \qquad\qquad \forall k \in [m], \forall 
i \in \{0, 1\}, \forall l \in [j-1] \label{eq:prev_x} \\
& \quad \text{Output:} \nonumber \\
& \qquad x_{k,i,j} \qquad\qquad \forall k \in [m], \forall i \in \{0,1\} \label{eq:x_online} \\
& \qquad st_j \label{eq:st_online} \\
& \text{Constraints:} \ ~\eqref{eq:u1} - \eqref{eq:st2} \nonumber
\end{align}
\end{subequations}

\section{Performance of Online Algorithm}
\label{sec:bounds}
This section examines the performance bounds of greedy online algorithms for the DVAMP problem. In this context, 'greedy' implies immediate deployment of a VM if NUMA node(s) are available, as online algorithms cannot determine if delaying deployment would be more beneficial. Let $OPT$ denote the $WT$ value for VM request set $V$ achieved by an optimal offline algorithm, and $ON$ represent the $WT$ value for $V$ of any greedy online algorithm. Given the possibility of $OPT=0$ and $ON>0$, the worst-case absolute optimality gap, defined as $\max_V\{ON-OPT\}$, is a more appropriate metric for performance analysis than the competitive ratio.

Several key variables are crucial in this analysis. First, $\mu = \max_{j} \{lt_j\} / \min_{j} \{lt_j\}$ represents the ratio of the maximum VM lifetime to the minimum. For simplicity, the minimum VM lifetime is assumed to be 1 and the maximum is $\mu$. Second, $TR = \sum_{d=1}^D \sum_{j=1}^n r^d_j * lt_j$ denotes the total time-resource of VM requests, serving as a descriptive metric for VM request characteristics. The worst-case optimality gap increases as a function of $TR$, making the ratio of $\max_V \{ON-OPT\}$ to $TR$ particularly significant. The analysis focuses on this ratio, which can be expressed as~\eqref{eq:ratio_definition}. Both variables are commonly employed in dynamic bin packing problems~\cite{coffman1983dynamic, ivkovic1998fully}.

\begin{equation}
\max_V\{\frac{ON-OPT}{TR}\}
\label{eq:ratio_definition}
\end{equation}

The analysis assumes $R_d=0.5$, implying a maximum resource capacity of 1 for all PMs. Additionally, the resource dimension $D$ is set to 1, which can be achieved when VM resource requirements are identical across all dimensions.

\begin{figure}[tb]
    \centering
    \includegraphics[width=1\columnwidth]{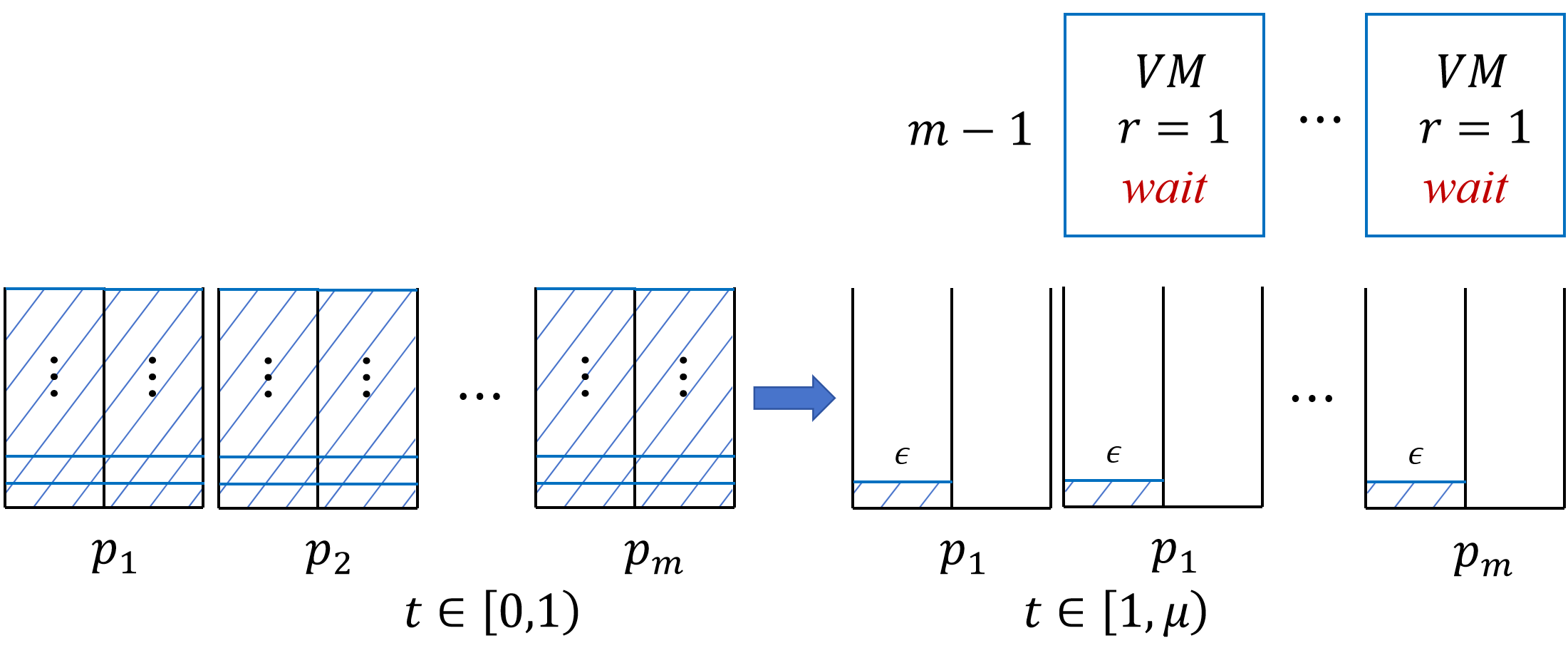}
    \caption{VM scheduling illustration for a greedy online algorithm. The algorithm initially deploys all VMs arriving at $t=0$, leading to $m-1$ VMs waiting when new VMs arrive at $t=1$.}
    \label{fig:greedy_online}
\end{figure}

Let $q$ be a large positive integer. Consider a VM request set comprising the following elements:
\begin{itemize}
    \item $m$ VMs with $r=\frac{1}{2q}, at=0, lt=\mu$;
    \item $2qm-m$ VMs with $r=\frac{1}{2q}, at=0, lt=1$;
    \item $m-1$ VMs with $r=1, at=1, lt=1$.
\end{itemize}
(For convenience, we omit the VM indices in this notation.)

Any greedy online algorithm will initially allocate all VMs arriving at $t=0$ to the cluster, with each NUMA node hosting $q$ VMs where $r=\frac{1}{2q}$. At $t=1$, $2qm-m$ VMs terminate, leaving $m$ VMs active. Due to the inherent limitations of online algorithms, it is impossible to predict which VMs will terminate at $t=1$. In the worst-case scenario, the $m$ active VMs occupy each PM, preventing the immediate deployment of the $m-1$ newly arriving VMs with $r=1$. These new VMs must wait until the initially active VMs terminate. Fig.~\ref{fig:greedy_online} illustrates this process.

The resulting total wait time is $ON=(m-1)(\mu-1)$, with $TR=m(\frac{\mu}{2q})+\frac{2qm-m}{2q}+(m-1)$. As $q$ approaches infinity:

\begin{equation}
\lim_{q\rightarrow+\infty} \frac{ON}{TR}=\frac{(m-1)}{(2m-1)} (\mu-1)
\label{eq:limit_q}
\end{equation}

\begin{figure}[tb]
    \centering
    \includegraphics[width=\columnwidth]{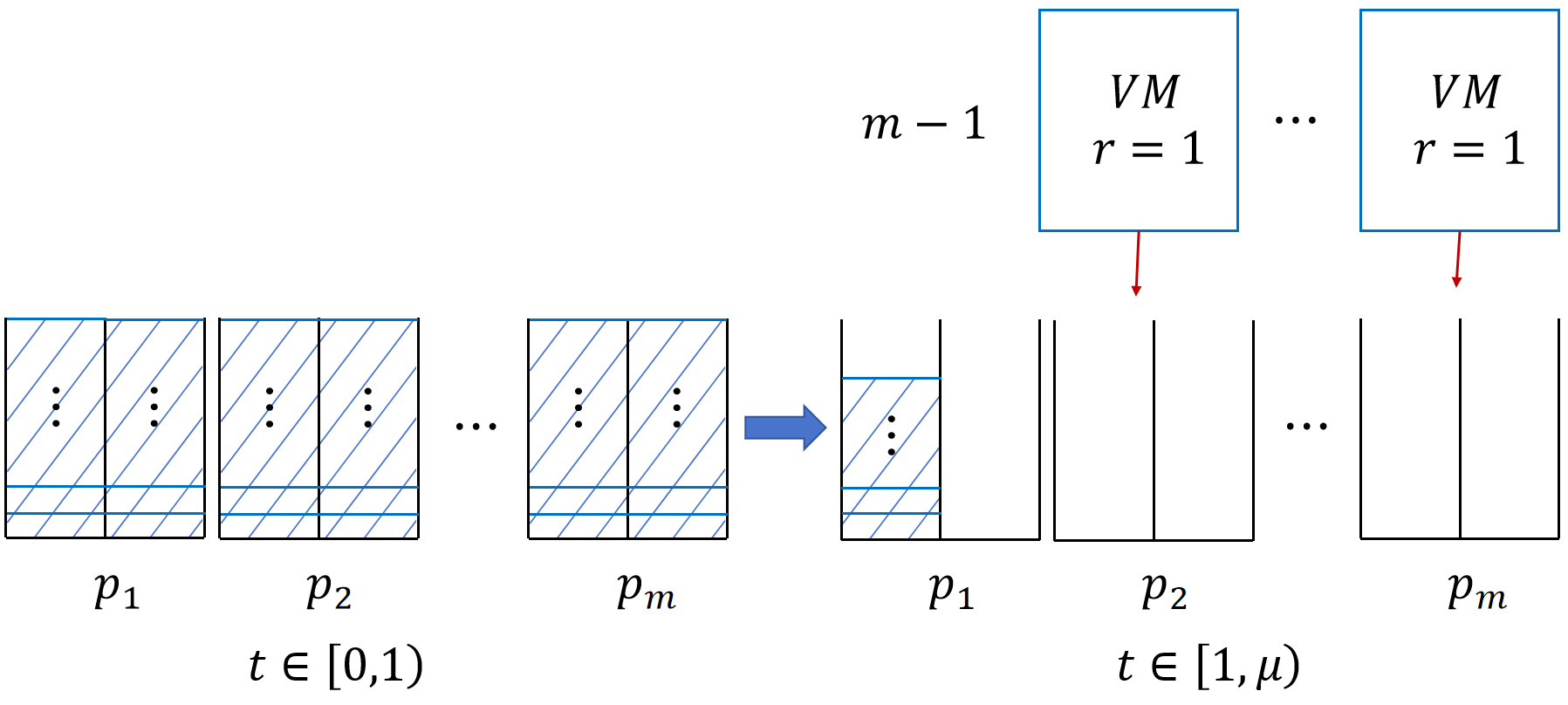}
    \caption{VM scheduling illustration for the optimal algorithm. By strategically placing VMs active after $t=1$ on the first NUMA node, the algorithm achieves immediate deployment of new VMs at $t=1$, minimizing wait time.}
    \label{fig:optimal}
\end{figure}

In contrast, an optimal algorithm can place the $m$ VMs that remain active after $t=1$ on the first NUMA node. This strategy allows immediate deployment of the $m-1$ VMs arriving at $t=1$ on the remaining $m-1$ PMs, eliminating wait time. Fig.~\ref{fig:optimal} depicts this optimal scheduling process.

In conclusion, for any greedy online algorithm, the worst-case performance relative to the optimal algorithm satisfies~\eqref{eq:worst_case_bound}.

\begin{equation}
\max_V \left \{ \frac{ON-OPT}{TR}\right \} \geq \frac{(m-1)}{(2m-1)} (\mu-1)
\label{eq:worst_case_bound}
\end{equation}

\section{Reinforcement Learning Algorithm}
\label{sec:RL}
\subsection{Reinforcement Learning Framework}
To address the complex optimization challenge of the DVAMP problem, we propose a reinforcement learning approach that can adapt to the dynamic nature of VM requests and resource utilization. We formulate the DVAMP problem as a Markov Decision Process (MDP) with partial observability, suitable for reinforcement learning techniques.

\textbf{Full State ($S_{\text{full}}$):} The full state of the system at time $t$ is represented as:
\begin{equation}
s^{\text{full}}(t) = (p_1(t), \cdots, p_m(t), \boldsymbol{r}_j, at_j, div_j)
\label{eq:full_state}
\end{equation}
where $p_k(t)$ represents the full information of all active VMs deployed in $p_k$ at time $t$. It is defined in the following equation:
\begin{equation}
p_k(t) = \{(st_l, i, \boldsymbol{r}_l) \mid x_{k,i,l}=1 \text{ and } \delta_l(t)=1\} 
\label{eq:pk_t}
\end{equation}
Here, $\boldsymbol{r}_l = (r^1_l, \cdots, r^D_l)$ is $v_l$'s resource array, and $v_j$ represents the next VM awaiting deployment. Specifically, the last three elements of~\eqref{eq:full_state} ($\boldsymbol{r}_j$, $at_j$, and $div_j$) represent the resource requirements, arrival time, and deployment strategy of the next VM to be scheduled. If no VM is pending, these three elements are set to null values.

\textbf{Observable State ($S_{\text{obs}}$):} To reduce complexity, the agent observes only the resource utilization of NUMA nodes:
\begin{equation}
s_j^{\text{obs}} = (\boldsymbol{u}_1(st_j), \cdots, \boldsymbol{u}_m(st_j), \boldsymbol{r}_j, st_j, at_j, div_j) 
\label{eq:obs_state}
\end{equation}
Here, $\boldsymbol{u}_k(st_j) = \{u^d_{k,i}(st_j) \mid \forall i \in \{0,1\}, \forall d \in [D]\}$ is the resource utilization of $p_k$ at $st_j$, calculated from active VM information. $st_j$ denotes the deployment time for $v_j$, which is either its arrival time or the earliest time when at least one NUMA node becomes available for deployment.

\textbf{Action ($A$):} Action $a_j \in [2m]$ determines the deployment location for $v_j$. If $div_j=0$, $v_j$ is deployed in the $\lfloor (a_j + 1) / 2 \rfloor$-th server and $[(a_j + 1)\mod 2]$-th NUMA node; otherwise, it is deployed across both NUMA nodes of the same server.

\textbf{Reward ($R$):} The reward $rew_j$ is defined as the negative wait time of \(v_j\). This formulation aligns with our objective to minimize total wait time. 
\begin{equation}
rew_j = R(s_j^{\text{obs}}, a_j) = -(st_j - at_j) 
\label{eq:reward}
\end{equation}

\textbf{Transition ($P$):} We assume discrete time steps, which aligns with the characteristics of real-world datasets such as Huawei-East-1 (to be described in detail in the experimental section). We define two types of transitions:
\begin{itemize}
    \item Action-based transition: $s^{\text{full}}(t) \times a_j \to s^{\text{full}}(t+1)$
    \item Automatic transition: $s^{\text{full}}(t) \to s^{\text{full}}(t+1)$
\end{itemize}
Action-based transitions occur when a VM is deployed, updating the VM installation status in the full state. The observable state $s_j^{\text{obs}}$ is then updated to reflect the changes in resource utilization. Automatic transitions reflect the termination of VMs that have completed their active duration.

\textbf{Policy ($\pi$):} The policy $\pi(a_j \mid s_j^{\text{obs}})$, denoted as $pr^{(a_j)}_j$, defines the probability of taking action $a_j$ given the observable state $s_j^{\text{obs}}$. For convenience, we define the policy vector as:
\begin{equation}
\boldsymbol{\pi}(s^{\text{obs}}_j) = (pr^{(1)}_j, \ldots, pr^{(2m)}_j)
\label{eq:policy_vector}
\end{equation}

This reinforcement learning framework provides a structured approach to addressing the DVAMP problem, balancing the trade-off between state observability and computational complexity while adapting to the dynamic nature of VM scheduling in cloud environments.

\subsection{Symmetries of DVAMP Problem}
The DVAMP problem exhibits inherent symmetries stemming from the equivalent status of PMs. Under arbitrary permutations of PM states, the policy should undergo a corresponding permutation, reflecting the interchangeable nature of PMs in the system. Specifically, when we apply any permutation to the PM states, the policy vector should be rearranged in the same manner, with the probabilities associated with each PM (represented by pairs of elements due to the two NUMA nodes per PM) following the same permutation.

To formally describe these symmetries, we employ the mathematical concept of transformation groups. A transformation on $s_j^{\text{obs}}$ can be represented as an element of the permutation group of elements $\boldsymbol{u}_1, \boldsymbol{u}_2, \ldots, \boldsymbol{u}_m$. We define the transformation group as $S_m'$, which is isomorphic to the $m$-element permutation group $S_m$. For any $\sigma = (\alpha_1, \ldots, \alpha_m) \in S_m$, where $\alpha_i$ represents the element that $i$ is mapped to, the corresponding element $L_\sigma \in S_m'$ is defined as:
\begin{equation}
L_\sigma[s_j^{\text{obs}}] = (\boldsymbol{u}_{\alpha_1}(st_j), \ldots, \boldsymbol{u}_{\alpha_m}(st_j), \boldsymbol{r}_j, st_j, at_j, div_j)
\label{eq:transformation_L}
\end{equation}
Similarly, the corresponding transformation on $\boldsymbol{\pi}$, denoted as $K_\sigma$, is defined such that $\forall k \in [m]$:
\begin{equation}
\begin{aligned}
K_\sigma[\boldsymbol{\pi}(s_j^{\text{obs}})]_{2k-1} & = pr_j^{(2\alpha_k - 1)} \\
K_\sigma[\boldsymbol{\pi}(s_j^{\text{obs}})]_{2k} & = pr_j^{(2\alpha_k)} 
\label{eq:transformation_K_L}
\end{aligned}
\end{equation}
where the subscripts \(2k-1,2k\) denote the element indices.

\begin{figure}[tb]
\centering
\includegraphics[width=\columnwidth]{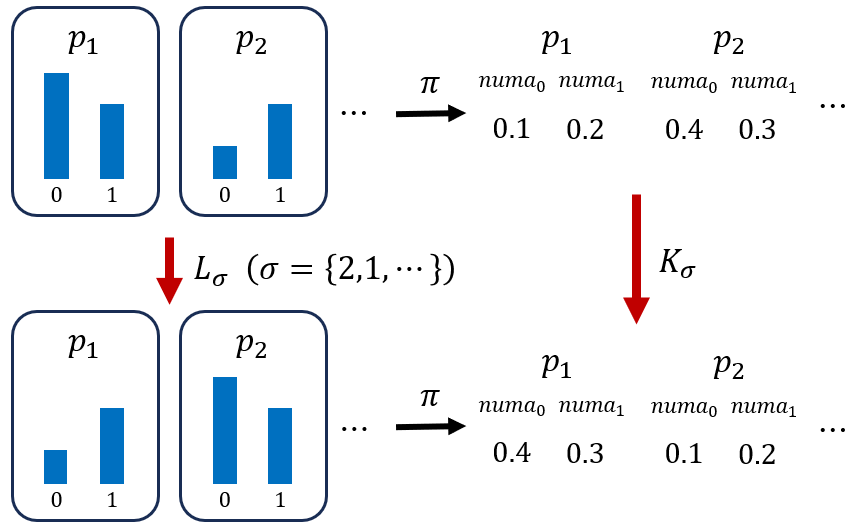}
\caption{Illustration of a symmetry-preserving policy in DVAMP. The upper row depicts the original resource utilization of NUMA nodes in $p_1$ and $p_2$, with corresponding policy probabilities. The lower row shows the scenario after swapping the states of $p_1$ and $p_2$. A symmetry-preserving policy ensures that the output probabilities for $p_1$ and $p_2$ are correspondingly swapped, maintaining consistency under PM permutations.}
\label{fig:sym_policy}
\end{figure}

We define a symmetry-preserving policy as one that respects the inherent symmetry of the DVAMP problem. Formally, a policy $\boldsymbol{\pi}(\cdot|\cdot)$ is considered symmetry-preserving if it satisfies:
\begin{equation}
K_\sigma[\boldsymbol{\pi}(s_j^{\text{obs}})] = \boldsymbol{\pi}(L_\sigma[s_j^{\text{obs}}])
\label{eq:sym_policy}
\end{equation}
Fig.~\ref{fig:sym_policy} illustrates this concept, demonstrating how a symmetry-preserving policy maintains consistency under various permutations of PM states.

\subsection{Symmetry Network Architecture}

Conventional deep reinforcement learning (DRL) typically employs convolutional neural networks (CNNs) or multilayer perceptrons (MLPs) to approximate Q-functions or policies, but these classical architectures fail to exploit the inherent symmetry in the DVAMP problem. We propose a novel network architecture that inherently satisfies the symmetry constraints, ensuring that the policy trained via DRL adheres to the symmetry condition defined in~\eqref{eq:sym_policy}. While we focus on deep Q-learning (DQN) for illustration due to its straightforward implementation, the proposed symmetry network is readily adaptable to other DRL methodologies.

Contemporary DQN implementations often utilize the dueling technique \cite{duelingDQN}, which decomposes the optimal Q-function into two components: the optimal state value function \(V(s_j^{\text{obs}})\) and the state-action advantage function \(A(a_j, s_j^{\text{obs}})\). This decomposition is expressed as \(Q(a_j, s_j^{\text{obs}}) = V(s_j^{\text{obs}}) + A(a_j, s_j^{\text{obs}})\). Post-training, the DQN policy is formulated as:
\begin{equation}
\pi(a_j \mid s_j^{\text{obs}}) = \begin{cases}
1, & \text{if } a_j = \arg\max_a Q(a, s_j^{\text{obs}}) \\
0, & \text{otherwise}
\end{cases}
\label{eq:dqn_policy}
\end{equation}

Given that the total cluster state should be invariant to PM ordering, a symmetry-preserving DQN policy must satisfy:
\begin{equation}
V(s_j^{\text{obs}}) = V(L_\sigma[s_j^{\text{obs}}]), \quad \forall L_\sigma \in S_m'
\label{eq:symmetry_v}
\end{equation}
For the advantage function \(\boldsymbol{A}(\cdot)\), we define an advantage vector \(\boldsymbol{A}(s_j^{\text{obs}}) = (A(1, s_j^{\text{obs}}), \ldots, A(2m, s_j^{\text{obs}}))\) that must satisfy:
\begin{equation}
K_\sigma[\boldsymbol{A}(s_j^{\text{obs}})] = \boldsymbol{A}(L_\sigma[s_j^{\text{obs}}]), \quad \forall L_\sigma \in S_m'
\label{eq:symmetry_a}
\end{equation}
This implies that \(A(2k-1, s_j^{\text{obs}})\) and \(A(2k, s_j^{\text{obs}})\) should depend only on \(p_k\) and inputs that are invariant to PM ordering.

\begin{figure}[tb]
\centering
\includegraphics[width=\columnwidth]{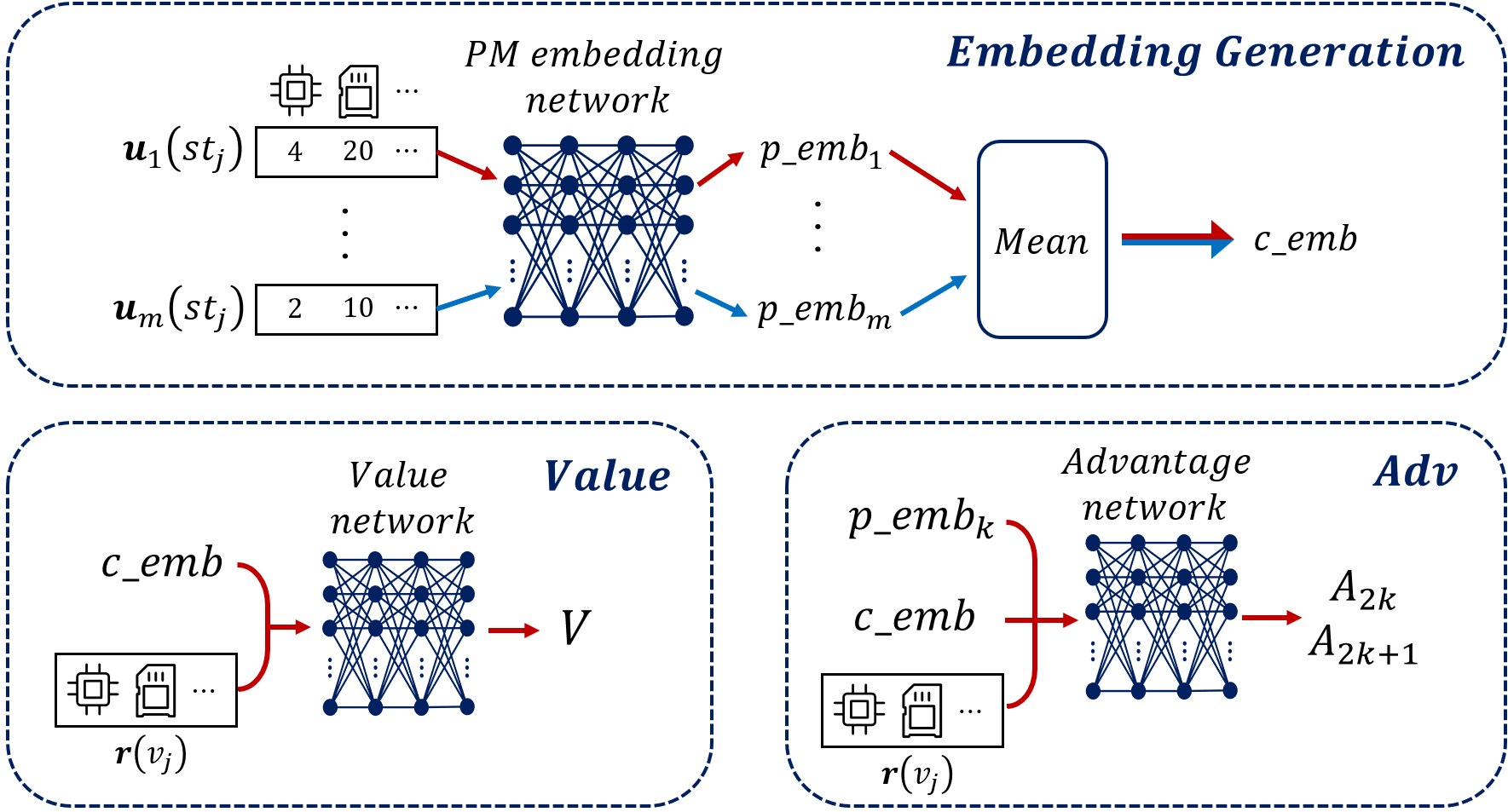}
\caption{Symmetry-preserving network architecture for DQN in the DVAMP problem.}
\label{fig:network}
\end{figure}

To achieve this symmetry-preserving architecture, we construct three key modules, as illustrated in Fig.~\ref{fig:network}.

\textbf{Embedding Generation Module:} This module constructs PM embeddings for each PM and a cluster embedding that is independent of PM ordering. It employs a shared PM Embedding Network across all PMs to generate individual PM embeddings, followed by computing their mean to obtain the cluster embedding. This mean aggregation technique eliminates PM ordering dependencies.

\textbf{Value Module:} In this module, we concatenate the cluster embedding with the resource requirements \(\boldsymbol{r}_j\) and process this through a value network to derive the state value.

\textbf{Advantage Module:} This module employs a common advantage network that takes as input the \(k\)-th server embedding, the cluster embedding, and the resource requirements \(\boldsymbol{r}_j\), outputting \(A(2k-1, s_j^{\text{obs}})\) and \(A(2k, s_j^{\text{obs}})\).

By combining~\eqref{eq:dqn_policy},~\eqref{eq:symmetry_v}, and~\eqref{eq:symmetry_a}, we can conclude that our DQN policy satisfies the required symmetry~\eqref{eq:sym_policy}. Consequently, our DQN method effectively leverages the symmetry inherent in the DVAMP problem, which is anticipated to yield superior performance, as demonstrated in Section~\ref{sec:experiment}.

\section{Performance Evaluation}
\label{sec:experiment}
\subsection{Experiment Settings}

\paragraph{Dataset, Environment}
Experiments use the Huawei-East-1 dataset~\cite{sheng_vmagent_2022}, containing 125,430 VM creation and 116,313 deletion requests over 30 days. As the DVAMP problem considers all requests to have active length time, only requests with deletion data are considered. The active length time is obtained by subtracting the deletion request time from the arrival time. Each request includes two resource requirements: CPU core and memory, with 15 types ranging from 1u1g to 64u128g.

For the cluster, similar to~\cite{sheng_vmagent_2022}, PM number $m=5$, the resource dimension $D=2$ which are cpu cores and memory, $R_1 = 40$, $R_2 = 90$, and $div_j = 1$ if $R_2 \geq 10$. For each episode, the environment (consisting of requests and PMs) is reset initially. Upon reset, cluster resource utilization is set to zero. The training set randomly selects initial request indices from the range 0 to 50,000. The validation and test sets randomly select from ranges 50,000 to 70,000 and 70,000 to 110,000, respectively. Both validation and test sets remain fixed throughout the experiments.

\paragraph{Compared Methods}
The proposed method (\textbf{SPANE-DQN}) is compared with several heuristic methods: First Fit, Balance Fit, DQN using MLP for approximate Q value (\textbf{MLP-DQN}), and MLP with data augmentation (\textbf{MLP-DQN-Aug}). The comparison with \cite{sheng_vmagent_2022} was not conducted due to differences in the considered objective and the lack of specification of its reward function, which is a key innovation point. Additionally, while \cite{mdp_homo} proposed a method to exploit arbitrary symmetry of MDP by limiting the linear space of the weight matrix from MLP, its implementation proves challenging due to the lack of support from common modules such as PyTorch for its training procedure.

\textbf{SPANE-DQN}: This method employs DQN with techniques such as dueling~\cite{duelingDQN}, double DQN~\cite{doubleDQN}, and $n$-step as the DRL method. It use the SPANE to approximate the Q functiom. The general training process involves:
\begin{enumerate}[leftmargin=*]
    \item Collecting data from 100 episodes without training; the data are in the form $(s_j^{\text{obs}}, a_j, rew_j, s_{(j+n)}^{\text{obj}})$ and are stored in replay memory, where $rew_j$ is the n-step reward.
    \item For each epoch, collecting data from one episode, then undergoing a learning procedure from the data in replay memory randomly.
    \item After a certain interval of epochs, performing validation and obtaining the validation loss. The model with the least validation loss is selected.
\end{enumerate}

\textbf{Balance Fit}: For VMs with $div_j = 1$, this method selects the first PM capable of deploying the VM. For VMs with $div_j = 0$, it first considers only PMs with at least one available NUMA node. Then, it chooses the PM with the largest NUMA nodes resource utilization difference, calculated by~\eqref{eq:balance_fit}:
\begin{equation}
\sum_{d=1}^D \frac{|u_{k,0}^d - u_{k,1}^d|}{R_d}
\label{eq:balance_fit}
\end{equation}
It then selects the less utilized NUMA node for VM deployment, aiming to balance the utilization load.

\textbf{MLP-DQN}: This method uses the same DRL method as Sym-DQN but employs MLP to approximate state value and state-action advantages.

\textbf{MLP-DQN-Aug}: In this method, after storing the original data, transformations are applied to enable the MLP to learn the symmetry of the original problem. Specifically, several random $L_{\sigma} \in S_m'$ are chosen, and the replay memory stores $(L_{\sigma}[s_j^{\text{obs}}], K_{\sigma}'[a_j], rew_j, L_{\sigma}[s_{(j+n)}^{\text{obj}}])$, where $K_{\sigma}'$ is defined by~\eqref{eq:k_sigma_prime} and~\eqref{eq:k_i}:
\begin{equation}
K_{\sigma}'[a_j] = 2k - i
\label{eq:k_sigma_prime}
\end{equation}
\begin{equation}
k = \sigma(\lfloor (a_j + 1) / 2 \rfloor), i = a_j \bmod 2
\label{eq:k_i}
\end{equation}
Here, $\lfloor a_j / 2 \rfloor$ and $i$ represent the corresponding server index and the NUMA node index, respectively. The transformation changes the server index from $\lfloor a_j / 2 \rfloor$ to $\sigma(\lfloor a_j / 2 \rfloor)$, retaining the NUMA index and the reward. The neural network structure for MLP-DQN-Aug is identical to that of MLP-DQN.

\paragraph{Hyperparameters}
Random grid search is employed to determine suitable hyperparameters for the DRL. The hyperparameters and environment parameters are shown in Table~\ref{tab:hyperparameters}. Best hyperparameters are selected based on the least validation loss. For SPANE-DQN, all three networks have only one hidden layer; the PM embedding network and value network hidden layer have a width of 8, while the advantage network has a width of 16. For both MLP-DQN and MLP-DQN-Aug, the network has two hidden layers with a width of 32. For MLP-DQN-Aug, after storing the original data, 23 random transformations are applied, and the transformed data are stored, resulting in each data point being augmented to 24 versions. To ensure a fair comparison, the training dataset size is kept the same for all DRL methods; hence, MLP-DQN-Aug collects data every 24 epochs.

\begin{table}[tb]
\centering
\caption{Hyperparameters and Environment Parameters}
\begin{tabular}{|c|c|}
\hline
\textbf{Parameter} & \textbf{Value} \\ \hline
$D, R_1, R_2$ & 2, 40, 90 \\ \hline
$d_{div}, c_{div}$ & 2, 10 \\ \hline
Activation function & ReLU \\ \hline
Optimizer & Adam \\ \hline
Epsilon-greedy probability & Linear decay from 0.6 to 0.0 \\ \hline
N step & 50 \\ \hline
$L_2$ weight decay & $10^{-8}$ \\ \hline
Learning rate & 0.01 \\ \hline
Epoch & 5000 \\ \hline
Batch Size & 1024 \\ \hline
Valid interval & 250 \\ \hline
Valid data size & 150 \\ \hline
Test data size & 1000 \\ \hline
\end{tabular}
\label{tab:hyperparameters}
\end{table}

\subsection{Wait Time Analysis}
\begin{figure}[tb]
    \centering
    \includegraphics[width=\columnwidth]{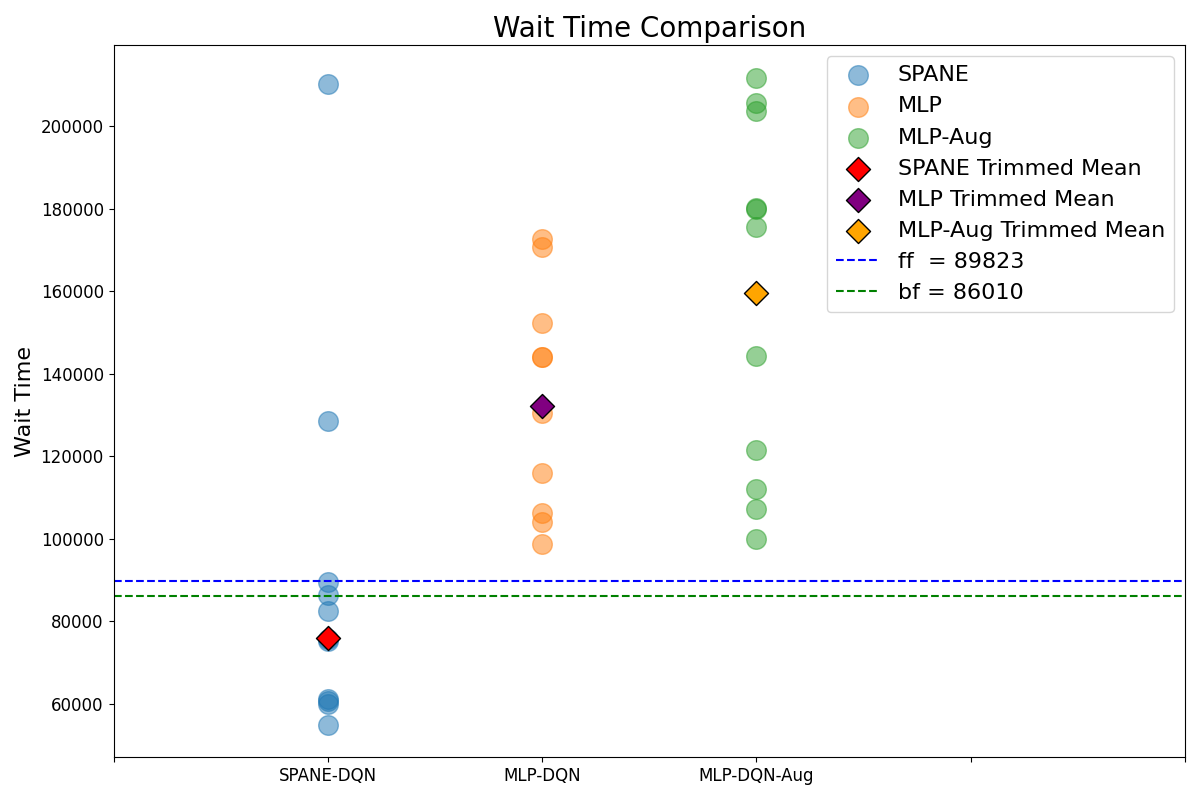}
    \caption{Wait time comparison across different methods. Circular markers represent mean wait times from 1000 test data points per experiment. The green and blue dashed lines indicate wait times for balance fit and first fit, respectively. Diamond markers denote the trimmed mean for each method.}
    \label{fig:wait_time_comparison}
\end{figure}

This section presents a comparative analysis of total wait times across heuristic methods, SPANE-DQN, MLP-DQN, and MLP-DQN-Aug. For each DQN method, 11 experiments were conducted, maintaining consistent parameters and validation sets while varying initialization and random training sets. Figure~\ref{fig:wait_time_comparison} illustrates the results.

\begin{table}[tb]
\centering
\caption{Total Wait Time for Different Methods}
\label{tab:WT}
\setlength{\tabcolsep}{12pt} % increase column spacing
\renewcommand{\arraystretch}{1.5} % increase row spacing
\begin{tabular}{|c|c|c|}
\hline
Method & \textbf{Trimmed Mean} & \textbf{Mean of Top 3} \\ \thickhline
SPANE-DQN        & \textbf{75,803}    & \textbf{63,697}    \\ \hline
MLP-DQN          & 132,144   & 117,552   \\ \hline
MLP-DQN-AUG      & 159,568   & 133,852  \\ \hline
First Fit        & 89,823    & 89,823  \\ \hline
Balance Fit      & 86,010    & 86,010  \\ \hline
\end{tabular}
\end{table}

Figure~\ref{fig:wait_time_comparison} reveals two notable outliers in the SPANE-DQN results. Consequently, a trimmed mean was calculated for each DQN method, excluding the top and bottom two results. We also calculate the mean of the three experiments with the lowest validation loss for pratical model selection. Table~\ref{tab:WT} presents these result, demonstrating that SPANE-DQN outperforms all other methods. Specifically, SPANE-DQN achieves wait times that are only 54\% of those obtained with MLP-DQN, indicating significantly higher efficiency. Additionally, selecting models with the lowest validation loss consistently yields better results.

The results indicate that both MLP-DQN and MLP-DQN-Aug perform significantly worse than the heuristic methods. This suggests that exploiting symmetry in an intelligent manner, as done in SPANE-DQN, can yield superior results. The poor performance of MLP-DQN-Aug is particularly noteworthy. While data augmentation may enable the DQN to learn the symmetry of the DVAMP problem, it potentially introduces high correlation within the replay memory data, which could adversely affect performance.

\subsection{Flexibility of SPANE}
The SPANE neural network architecture demonstrates remarkable flexibility, as it is independent of the PM cluster size. This characteristic allows the SPANE-DQN model to schedule VM requests without retraining when the number of PMs changes. To evaluate this flexibility, we conducted experiments using the top three SPANE-DQN models trained with $m=5$ PMs. We then compared the total wait time of these models to that of the best heuristic method, Balance Fit, across varying numbers of PMs.

\begin{figure}[tb]
    \centering
    \includegraphics[width=\columnwidth]{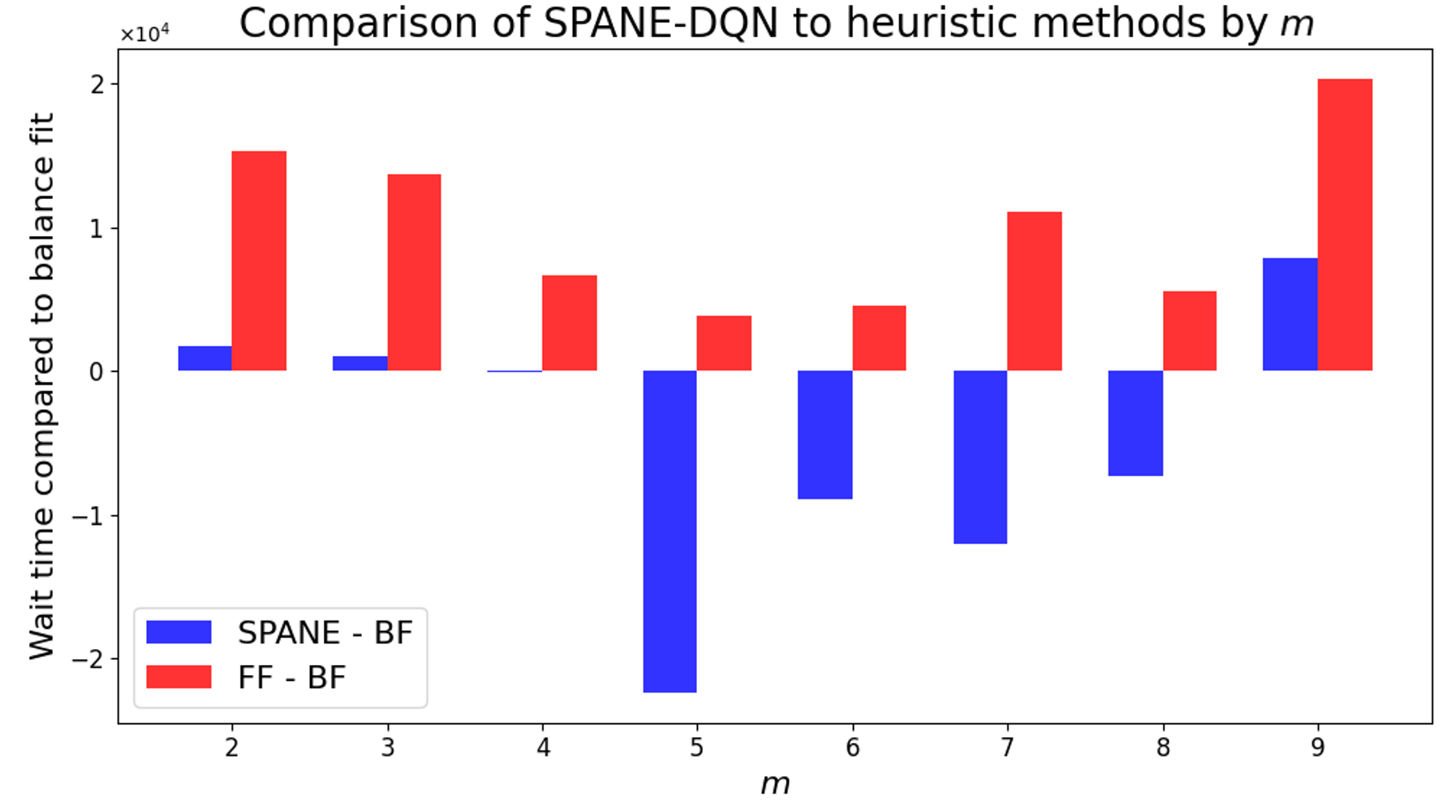}
    \caption{Comparison of SPANE-DQN to heuristic methods for different numbers of PMs ($m$). The graph shows the difference in wait time between SPANE-DQN and Balance Fit (blue bars), and between First Fit and Balance Fit (red bars). Negative values indicate better performance than Balance Fit.}
    \label{fig:spane_flexibility}
\end{figure}

Figure~\ref{fig:spane_flexibility} illustrates the comparative performance of SPANE-DQN and First Fit against Balance Fit for various PM counts ($m$). The y-axis represents the difference in wait time, with negative values indicating superior performance compared to Balance Fit.

The results reveal important insights about SPANE-DQN's performance and flexibility. SPANE-DQN achieves its best performance at $m=5$, aligning with the training configuration and underscoring the effectiveness of the learning process. More importantly, the graph demonstrates SPANE-DQN's flexibility across different cluster sizes. For $m=6, 7,$ and $8$, SPANE-DQN consistently outperforms Balance Fit, despite being trained on a different PM count. This adaptability indicates that the learned scheduling strategies can generalize to larger cluster sizes without retraining.

At $m=4$, SPANE-DQN's performance is slightly inferior to Balance Fit, though the difference is marginal. This suggests that the model maintains relatively competitive performance even when applied to cluster sizes smaller than its training configuration. The performance variation across different $m$ values hints at potential differences in optimal scheduling strategies for varying cluster sizes. For larger values of $m$ ($5$ and above), SPANE-DQN generally outperforms Balance Fit, which may indicate that the model's learned strategies are particularly effective for these cluster sizes.

These observations raise interesting questions about the relationship between cluster size and optimal scheduling strategies. It's possible that as cluster size increases, there might be some convergence in effective scheduling approaches, which could potentially benefit methods like SPANE-DQN that are designed to capture complex patterns. For smaller clusters, the performance differences suggest that optimal strategies might vary more significantly, presenting an opportunity for further investigation and potential refinement of the SPANE-DQN model for these scenarios.

The consistent outperformance of both SPANE-DQN and Balance Fit over First Fit across all $m$ values emphasizes the value of sophisticated scheduling methods in VM allocation tasks, highlighting the limitations of simpler heuristics in complex scenarios.

These findings underscore the flexibility and effectiveness of the SPANE-DQN approach in VM scheduling across various cluster sizes. The ability to maintain and often improve performance without retraining represents a significant advantage in terms of computational efficiency and adaptability to changing infrastructure requirements.
\balance

\section{Related Work} 
\subsection{Dynamic VM Scheduling Problem}
The dynamic VM Scheduling (DVMS) Problem is a specific application of the dynamic bin packing (DBP) in the cloud computing environment. There has been a large amount of prior work focusing on DVMS with different objective functions. Some works focus on minimizing the number of hosts \cite{chan2009dynamic, murhekar2023dynamic}, which is the same as the DBP. A large amount of research works have been done to analyze the competitive ratios of various algorithms for DBP. Coffman et al. \cite{coffman1983dynamic} proved that First Fit algorithm has competitive ratios between 2.75 and 2.897. Ivkovic et al. \cite{ivkovic1998fully} generalized the problem by considering that packed items can be moved to different bins. 
Some works focus on minimizing the usage time of the hosts \cite{KongZSG23,tang2016first,dynamicWithpre-2023}. Tang et al.\cite{tang2016first} proposed an algorithm to minimize the host run time for DVMS. Liu et al. \cite{dynamicWithpre-2023} predicted the duration for each item and develop an online algorithm with consistency and robustness. Some works consider different server architecture and maximize the number of created VMs during scheduling \cite{sheng_learning_2022,sheng_vmagent_2022}. Sheng et al. \cite{sheng_learning_2022} considered the multi-NUMA architecture and formulated it into a reinforcement learning framework. Then, an efficient RL-based algorithm is proposed to maximize the number of scheduled VMs. In contrast with these work, our work treats the VM waiting time as objective and analyze the competitive ratio under multi-NUMA scenario. 
% NUMA should be decleared earlier
% RL should be decleared earlier
% Aniket et al. \cite{murhekar2023dynamic}

\subsection{Reinforcement Learning for Scheduling}
Cloud computing environments are becoming increasingly complex and volatile due to unpredictable user behaviors, heterogeneous servers, and a high volume of requests. To effectively schedule resource in the cloud, deep reinforcement learning has demonstrated considerable advantages in many complex scheduling scenarios \cite{wang2021energy,siddesha2022novel,zhou2024deep,peng2015random,cheng2018drl,cheng2022cost}. Wang et al. \cite{wang2021energy} optimized resource scheduling based on QoS feature learning and improved Q-learning algorithm. Chen et al. \cite{chen2020resource} proposed a prediction-based Q-learning algorithm to adaptively allocate resources for dynamic cloud workloads. Siddesha et al. \cite{siddesha2022novel} handled different sizes of tasks and varied number of VMs with deep RL. However, the state space and the action space for these problems are extremely huge, which leads to poor convergence and a large overhead. Most works also assume that the VM last time is known in advance. In this work, we focus on the dynamic scheduling problem (the VM last time is unknown until it is explicitly stopped by the user) and consider the multi-NUMA server architecture, which makes the state and action space more complex. 

\section{Conclusion}
We introduces a novel variation of the dynamic vector bin-packing problem, referred to as DVAMP, within the context of multi-NUMA systems. We formally define the problem, providing MILP formulations for both offline and online versions, and derive a performance bound for the greedy online algorithm. To address the challenges posed by DVAMP, we propose SPANE, a symmetry-preserving deep reinforcement learning framework that ensures allocation results are invariant to permutations of physical machine states, thereby improving both learning efficiency and solution quality. Extensive experiments on real-world cloud datasets demonstrate that our proposed RL algorithm significantly outperforms existing baselines, achieving notable reductions in VM wait time.

\section*{Acknowledgment}
This work was supported by the National Key R\&D Program of China [2024YFF0617700], the National Natural Science Foundation of China [U23A20309, 62172276, 62372296, 62272302], Shanghai Municipal Science and Technology Major Project [2021SHZDZX0102], and the Huawei Cloud [TC20230526044].

\newpage
\balance
\bibliographystyle{IEEEtran} 
\bibliography{references}
\end{document}